\documentclass[11pt]{article}

\usepackage{acl}
\usepackage{times}
\usepackage{latexsym}
\usepackage{amsmath,amssymb,mathtools}
\usepackage{booktabs}
\usepackage[T1]{fontenc}
\usepackage[utf8]{inputenc}

\usepackage{microtype}
\usepackage{verbatim}
\usepackage{subcaption}

\usepackage{inconsolata}

\usepackage{graphicx}
\usepackage{xurl}

\title{Are LLMs Good Financial User Simulators? Multi-view Investor Logic Alignment (MILA)}

\author{
  Jiajie He\textsuperscript{1} \quad
  Yuanhong Jiang\textsuperscript{1} \quad
  Xintong Chen\textsuperscript{3} \quad
  Dongling Ni\textsuperscript{1} \quad
  Wenjin Liu\textsuperscript{2} \quad \\[0.5em]
  \textsuperscript{1}Hithink Research \quad
  \textsuperscript{2}Nanyang Technological University \quad
  \textsuperscript{3}McMaster University \\ [0.5em]
  \texttt{jiajiehe@myhexin.com, yuanhongjiang@myhexin.com, chen54xt@mcmaster.ca} \\
  \texttt{donglingni@myhexin.com,wenjinliu23@outlook.com}
}

\begin{document}

\maketitle 

\begin{abstract}
Large language models (LLMs) are increasingly used as user simulators, yet it remains unclear whether their predictions faithfully reproduce the evolving decisions of individual users. We investigate this question in a controlled longitudinal paper-trading study with 80 participants, where user interactions, simulated transactions, virtual portfolio states, and point-in-time market information are aligned under a rolling next-day prediction protocol. We evaluate behavioral fidelity hierarchically, from trade occurrence to action structure, asset selection, and downstream portfolio consequences.

Across 1,239 aligned user-days, no evaluated LLM reliably outperforms a simple recent-activity persistence baseline for predicting whether a user trades. Fidelity further deteriorates at finer levels: models struggle to recover buy--sell structure and traded assets, and similar activity-level predictions can lead to substantially different portfolio trajectories. Controlled evidence ablations show that recent trading history strongly governs activity prediction, whereas asset selection is substantially more sensitive to the available evidence. An observational analysis further finds that intensified ticker-specific research predicts imminent trading, but diagnostic tests do not support a causal interpretation. These findings suggest that current LLMs capture useful short-term behavioral regularities without yet recovering a stable individual decision mechanism.
\end{abstract}

\section{Introduction}

Large language models (LLMs) are increasingly used to simulate human behavior in dialogue, recommendation, and interactive decision-making~\cite{chan2023chateval,chen2026realworldhumanbehaviorsimulation,10.1145/3586183.3606763,ren-etal-2024-bases}. Their ability to integrate heterogeneous contextual information and generate flexible, context-dependent actions makes them a natural foundation for user simulation. However, generating plausible behavior is not necessarily equivalent to faithfully reproducing the behavior of a particular individual. A simulator may match aggregate behavioral statistics while acting at the wrong times, selecting different objects, or responding to different evidence than the user it is intended to represent.

This distinction becomes especially important in longitudinal decision-making. User actions are not independent observations: each decision is conditioned on prior behavior, the current state of the environment, and the consequences of earlier actions. Financial decision-making provides a useful setting for studying this problem because these dependencies are explicit. Whether an investor trades may depend on recent activity, while the direction and asset involved may additionally depend on current holdings, asset-specific attention, and changing market conditions. A prediction can therefore appear correct at one level while remaining behaviorally inconsistent at another.

Existing financial-agent research has primarily evaluated trading performance, recommendation quality, or aggregate market behavior~\cite{repec:arx:papers:2407.18957,NEURIPS2025_5bf234ec,10.1145/3805712.3808658}. Recent work has also explored LLM-based investor simulation and evolving financial-user states~\cite{gong2026shijianbenchdialoguedecisionlonghorizon}. These studies establish the potential of LLMs for modeling financial interactions, but leave a more basic question comparatively underexplored: \emph{when an LLM produces behavior similar to an individual user, what aspects of that user's decision process have actually been reproduced?}

We study this question through a controlled longitudinal paper-trading setting. Eighty participants interact with real-time market information and make simulated investment decisions using non-redeemable virtual funds. The resulting trajectories contain cross-scenario platform interactions, executed paper trades, virtual portfolio states, and contemporaneous market information. At each prediction cutoff, the simulator receives only information observable before that time and predicts the participant's behavior on the next trading day. This rolling design allows us to evaluate temporally aligned individual decisions rather than aggregate behavioral similarity alone.

Our analysis separates behavioral fidelity into increasingly demanding levels. We first ask whether the simulator predicts \emph{whether} an individual trades. Conditional on trading, we examine whether it recovers the structure of the action, including buying, selling, or both; which assets are involved; and whether the resulting simulated decisions induce a similar portfolio trajectory. This decomposition is important because success at a coarse level need not imply fidelity at a finer one. A model may reproduce a user's overall trading frequency while trading on different days, correctly identify a trading day while selecting the wrong assets, or achieve moderate action agreement while producing a substantially different economic trajectory.

We further distinguish \emph{behavioral outcome fidelity} from \emph{decision-process fidelity}. Correct prediction alone does not establish that the simulator relies on the same information as the user. We therefore intervene on the simulator's input by withholding recent trading history, portfolio state, recent platform behavior, market context, and long-horizon profile information. These controlled ablations reveal which evidence sources causally affect the model's predictions. We complement this analysis with an observational study of real participant behavior, examining whether intensified ticker-specific research precedes subsequent trading. Importantly, we distinguish evidence that changes the simulator's prediction from evidence that is associated with user behavior, and both from evidence that can be identified as causally affecting the user's decision.

Our results reveal a substantial gap between coarse behavioral similarity and individual decision fidelity. For trade occurrence, a simple recent-activity persistence rule achieves performance competitive with or stronger than the evaluated LLMs, indicating that apparently strong next-action prediction can largely reflect short-term behavioral momentum. Fidelity further deteriorates when recovering buy--sell structure and traded assets. Moreover, similar activity-level predictions do not necessarily produce similar portfolio trajectories, demonstrating that agreement on \emph{whether} a user trades can obscure important disagreement about \emph{what} the user does.

The evidence analysis provides a complementary explanation. Removing recent trading history causes a large degradation in trade-occurrence prediction, whereas removing other evidence sources has comparatively limited effects on the binary activity decision. Asset predictions, in contrast, change substantially when almost any evidence group is withheld, suggesting a less stable asset-selection mechanism. In the observed participant trajectories, intensified ticker-specific research is positively associated with imminent same-ticker trading, but residual imbalance and placebo tests prevent a causal interpretation. Together, these results suggest that current LLM simulators capture predictive regularities in investor behavior more reliably than they recover stable, asset-specific, or causal decision mechanisms.

\paragraph{Contributions.}
Our contributions are threefold:
\begin{itemize}
\item We provide a longitudinal empirical study of LLM-based financial user simulation under strict point-in-time evaluation, separating behavioral fidelity across trade occurrence, action structure, asset selection, and downstream portfolio consequences.

\item We identify a hierarchy of behavioral fidelity: current LLMs can reproduce aspects of coarse trading behavior, but agreement progressively weakens for temporally aligned decisions, buy--sell structure, asset selection, and induced portfolio trajectories. We further show that a simple recent-activity persistence rule remains a strong baseline for trade-occurrence prediction.

\item We distinguish predictive dependence from behavioral mechanism through controlled evidence interventions and observational analysis. Recent trading history strongly governs the simulator's activity prediction, whereas asset-level predictions depend on a less stable combination of evidence; meanwhile, pre-trade research activity predicts subsequent trading without supporting causal identification.

\end{itemize}

\section{Task Formulation}
\label{sec:task_formulation}

\textbf{Rolling Simulation Setting}: Let $i$ index investors and $t$ index U.S.\ equity trading days. For each target trading day $t+1$, we define the prediction cutoff $c_t$ as the market close on day $t$. The simulator is restricted to point-in-time information observable at or before $c_t$:

\begin{equation}
\label{eq:inputs}
\mathcal{I}_{i,t}
=
\left(
\mathbf{u}_{i,t},
\mathbf{h}_{i,\leq t-7},
\mathbf{r}_{i,t-6:t},
\mathbf{p}_{i,t},
\mathbf{m}_{t}
\right),
\end{equation}

where $\mathbf{u}_{i,t}$ denotes the investor profile constructed from expanding historical records; $\mathbf{h}_{i,\leq t-7}$ denotes long-term executed trading history preceding the recent window; $\mathbf{r}_{i,t-6:t}$ denotes cross-scenario behavior during the most recent seven trading days; $\mathbf{p}_{i,t}$ denotes the portfolio state at the cutoff; and $\mathbf{m}_{t}$ denotes market information observable through day $t$. Recent behavior includes transactions, stock and portfolio views, news consumption, and other in-platform interactions that may indicate emerging investment intent.

Given this information, the simulator predicts the investor's structured action on the
next trading day:

\begin{equation}
\label{eq:prediction}
\hat{\mathbf{a}}_{i,t+1}
=
f_{\theta}\left(\mathcal{I}_{i,t}\right).
\end{equation}

\begin{figure*}[t]
  \centering
  \includegraphics[width=\textwidth]{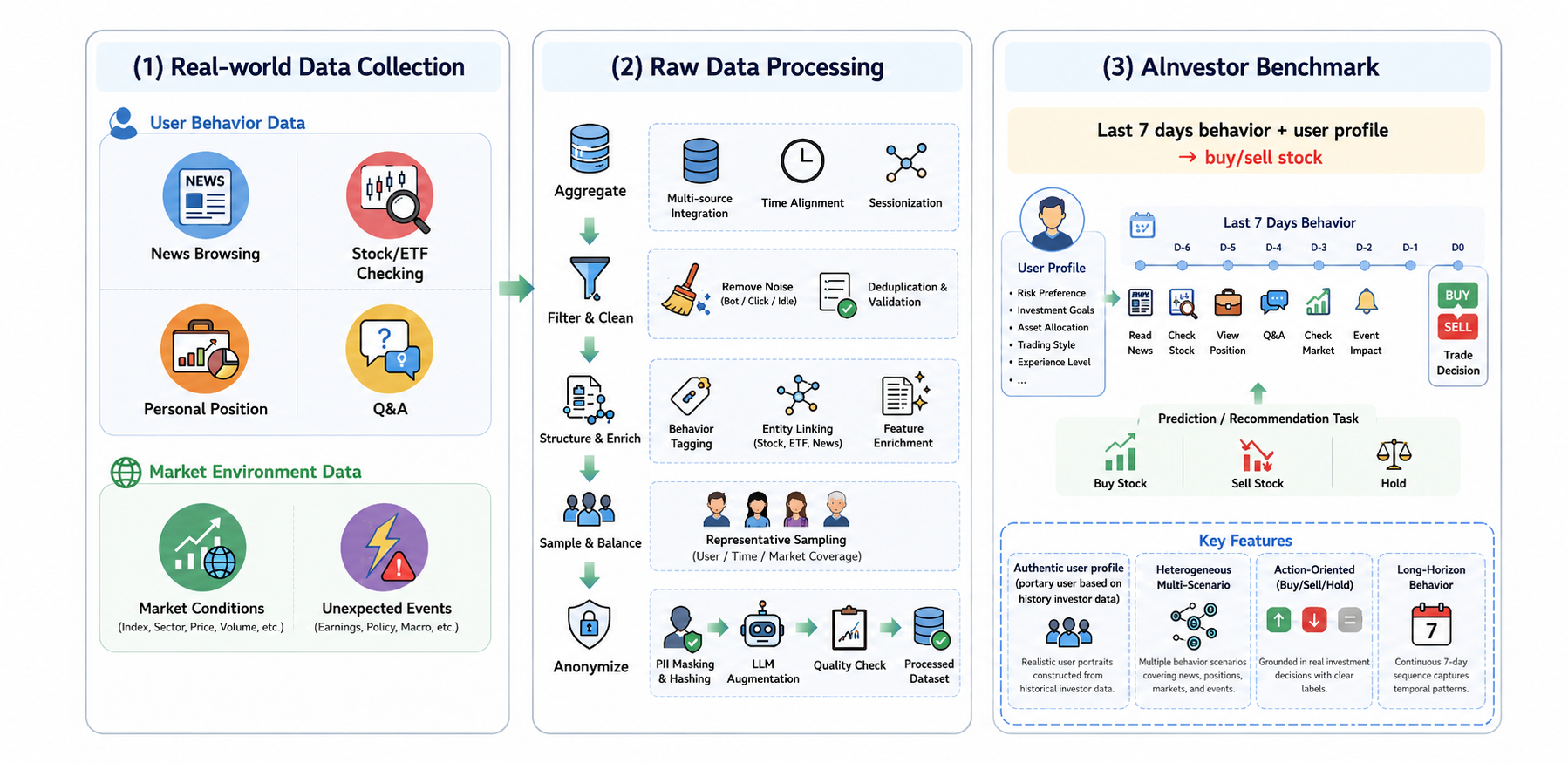}
  \caption{
  Overview of the rolling investor-simulation framework. At each cutoff $c_t$, the
  simulator constructs a point-in-time investor representation from the expanding
  profile, long-term trading history, recent seven-trading-day behavior, current
  portfolio state, and contemporaneous market context. It then predicts the investor's
  structured action on the next trading day $t+1$. All inputs are reconstructed using
  only information observable at or before the cutoff.
  }
  \label{fig:pipeline_user_prediction}
\end{figure*}

All time-varying inputs are reconstructed independently at each cutoff to ensure point-in-time validity and prevent future information leakage. During evidence-ablation
experiments, derived representations, including the investor profile, are recomputed
after the relevant source is removed to prevent information from leaking across evidence
groups. Figure~\ref{fig:pipeline_user_prediction} illustrates the complete rolling
simulation framework.

\section{Task Evaluation}
\label{sec:task_Evaluation}

We formulate personalized investor simulation as a rolling-origin, hierarchical prediction problem. At each prediction cutoff $t$, the simulator observes only information available before $t$ and predicts a structured sequence of investor decisions: whether the investor will trade, the trade direction, the assets involved, the position-level changes, and the transaction scale. These decisions jointly determine the predicted portfolio transition.

Our objective is \emph{behavioral fidelity} rather than portfolio optimization. Unlike trading agents, which are commonly evaluated by realized returns or recommendation quality, a personalized investor simulator should reproduce the decisions of the particular investor it represents---including the decision to abstain from trading---even when those decisions are not
financially optimal.

We evaluate fidelity along two complementary dimensions. First, \emph{behavioral outcome fidelity} measures how accurately the simulator predicts the investor's realized actions under strictly temporal evaluation. Second, \emph{decision-process fidelity} examines whether the simulator recovers an investor-specific decision policy rather than relying primarily on population-level trading frequencies or market regularities. We operationalize decision-process fidelity by testing whether predictions respond to investor-specific evidence, including historical behavior, portfolio state, research activity, and market context, and whether these sensitivities are consistent with the conditional patterns observed in the investor's actual decisions. We assess such sensitivity through controlled input ablations and counterfactual interventions rather than relying on the model's self-generated explanations.

\subsection{Hierarchical prediction of investor actions.}
We formulate action simulation as a five-stage hierarchical prediction task. Let the realized executions on day $t+1$ be

\begin{equation}
\mathcal{L}_{i,t+1}
=
\left\{
(j_k,\sigma_k,n_k)
\right\}_{k=1}^{K_{i,t+1}},
\end{equation}

where $j_k$ is the traded asset, $\sigma_k\in\{+1,-1\}$ indicates a buy or sell, and $n_k$ is the executed notional. First, the simulator predicts whether the investor trades:

\begin{equation}
y^{\mathrm{trade}}_{i,t+1}
=
\mathbb{I}\left[K_{i,t+1}>0\right].
\end{equation}

Conditional on trading, it predicts the day-level action composition:

\begin{equation}
y^{\mathrm{dir}}_{i,t+1}
\in
\{\textsc{buy},\textsc{sell},\textsc{buy\&sell}\}.
\end{equation}

The \textsc{buy\&sell} label represents portfolio rebalancing in which both buy and sell transactions occur on the same day. The simulator then ranks the eligible assets and returns the top-$k$ assets most likely to be traded. For each selected asset, it predicts the corresponding
position transition:

\begin{equation}
y^{\mathrm{pos}}_{i,t+1,j}
\in
\{\textsc{start},\textsc{add},\textsc{trim},\textsc{exit}\}.
\end{equation}

Here, \textsc{start} initiates a new position, \textsc{add} increases an existing position, \textsc{trim} reduces but retains a position, and \textsc{exit} closes it. Finally, the simulator predicts the normalized signed trade scale

\begin{equation}
\omega_{i,t+1,j}
=
\frac{
\sum_k \sigma_k n_k\mathbb{I}[j_k=j]
}{
V_{i,t}
},
\end{equation}

where $V_{i,t}$ is the portfolio value at the cutoff. These predictions jointly determine the simulated transaction-induced portfolio transition. The hierarchical prediction process is summarized as

\begin{equation}
\label{eq:prediction_hierarchy}
y^{\mathrm{trade}}
\rightarrow
y^{\mathrm{dir}}
\rightarrow
\mathcal{S}
\rightarrow
\mathbf{y}^{\mathrm{pos}}
\rightarrow
\boldsymbol{\omega}.
\end{equation}

Here, the stages correspond to trade occurrence, trade direction, asset selection, position transition, and trade scale, respectively. Portfolio transition is treated as an end-to-end outcome derived from these five components rather than as an additional prediction stage.

\paragraph{Evaluation.}
We evaluate trade occurrence, trade direction, and position transition using accuracy and macro-F1\cite{Takahashi2022}. Asset selection is evaluated using Hit@$5$\cite{10.1145/3038912.3052569,10.1145/3705328.3748052} and NDCG@$5$\cite{10.1145/582415.582418}. We report both stage-wise results, using ground-truth outputs from preceding stages, and end-to-end results based entirely on the simulator's predictions. Asset selection is performed over a point-in-time candidate set $\mathcal{C}_{i,t}$ containing current holdings, previously attended assets, and market-matched distractors. Target-day assets are not added based on information observed after cutoff $c_t$.

\subsection{Evaluating investor-specific decision policies.} Correctly predicting an action does not necessarily imply that the simulator has recovered the investor's decision policy. It may instead exploit population-level trading frequencies, asset popularity, or aggregate market regularities. We therefore evaluate whether predictions depend on investor-specific and decision-relevant evidence.

We organize the model input into five evidence groups: investor profile,
long-term trading history, recent behavior, portfolio state, and market
context. For each group $g$, we compare the original prediction with a controlled leave-one-group-out prediction:

\begin{equation}
\Delta_g
=
D\left(
f_{\theta}(\mathcal{I}_{i,t}),
f_{\theta}(\mathcal{I}^{(-g)}_{i,t})
\right),
\end{equation}

where $\mathcal{I}^{(-g)}_{i,t}$ removes evidence group $g$ and $D(\cdot,\cdot)$ measures changes in predicted probabilities, ranked assets, or discrete actions. We report probability shifts, action-flip rates, ranking changes, and performance degradation under each intervention.

These experiments identify which evidence affects the simulator's outputs. They do not, however, establish that the same evidence causally determines the real investor's behavior. We therefore refer to them as
\emph{evidence-sensitivity probes} rather than causal identification.

\paragraph{Evaluation.}
For each prediction, the simulator generates a structured natural-language explanation that links the predicted action to evidence available before the cutoff. The Sample is shown in the Appendix\ref{sec:human_evaluation}. We evaluate explanations from two complementary perspectives. First, human annotators assess\cite{deyoung-etal-2020-eraser}:

\begin{enumerate}
  \item \textbf{Evidence grounding}: whether the cited evidence is present in the pre-cutoff record;
  \item \textbf{Investor specificity}: whether the explanation reflects the focal investor rather than generic financial reasoning;
  \item \textbf{Decision consistency}: whether the evidence coherently
  supports the predicted action;
  \item \textbf{Historical plausibility}: whether the explanation is
  compatible with the investor's previously observed decisions.
\end{enumerate}

Second, we evaluate whether the model actually relies on the evidence cited in its explanation. For a cited evidence group $g$ and predicted action $\hat{a}$, we define interventional faithfulness as

\begin{equation}
F_g
=
s_{\theta}
\left(
\hat{a}\mid\mathcal{I}_{i,t}
\right)
-
s_{\theta}
\left(
\hat{a}\mid\mathcal{I}^{(-g)}_{i,t}
\right),
\end{equation}

where $s_{\theta}$ is the model score assigned to $\hat{a}$. If calibrated scores are unavailable, we instead report whether removing the cited evidence changes the predicted action or asset ranking. Human evaluation measures whether an explanation is grounded, specific, and plausible, whereas interventional faithfulness measures whether the prediction actually depends on the cited evidence. Neither alone demonstrates access to the investor's unobserved psychological process. 

\section{Experiments}
\label{sec:experiments}

We treat the current study as a pilot evaluation aimed at characterizing the
feasibility and failure modes of LLM-based financial user simulation rather
than establishing population-level performance. Experimental details are
provided in Appendix~\ref{sec:experiment_setting}.

\subsection{RQ1: How Faithful Are LLMs as Financial User Simulators?}
\label{sec:rq1}

We evaluate simulator fidelity hierarchically. A faithful financial user
simulator must reproduce not only \emph{whether} a user trades, but also the
structure of the action, the assets involved, and ultimately the economic
consequences of those decisions. We therefore organize the evaluation as

\[
\resizebox{0.95\columnwidth}{!}{$
\text{trade occurrence}
\rightarrow
\text{side structure}
\rightarrow
\text{asset selection}
\rightarrow
\text{economic consequence}
$}
\]

This decomposition is important because agreement at an earlier layer does not
necessarily imply fidelity at the next.

\textbf{LLMs capture trading propensity, but a simple persistence rule remains
competitive.}
Across 1,239 confirmed user-days, 307 contain an observed trade and 932 contain
no trade, corresponding to a real-user trade rate of 24.78\%. The simulators
exhibit markedly different activity biases. Gemini-3-Flash-preview trades much
more frequently than the observed users, whereas GPT-5.6-Sol is substantially
more conservative. GPT-5.6-Luna nearly reproduces the aggregate trade rate, but
aggregate agreement does not imply that the model trades on the correct dates.

This distinction is reflected in Table~\ref{tab:trade_no_trade}. GPT-5.6-Sol
achieves the highest LLM accuracy, but does so with low recall of actual trade
days. Gemini-3-Flash-preview shows the opposite behavior, recovering many more
real trades at the cost of substantially more false positives. More
importantly, the recent-activity persistence baseline---which predicts a trade
when the preceding trading day contains an execution---achieves 79.82\%
accuracy and 0.714 macro-F1, exceeding all LLM point estimates. Participant-level
bootstrap comparisons do not establish a reliable improvement of any LLM over
this simple behavioral rule.

The result suggests that a substantial component of next-day trading behavior
can already be explained by short-term activity persistence. A simulator may
therefore obtain apparently strong activity accuracy without recovering a
richer model of the user's investment decision process.

\begin{table*}[t]
\centering
\small
\caption{Trade-occurrence prediction (\%) on 307 confirmed trade days and
932 confirmed no-trade days. The recent-activity persistence baseline predicts
a trade when the preceding trading day contains a recorded execution. No LLM
reliably improves over this simple behavioral baseline.}
\label{tab:trade_no_trade}
\begin{tabular}{lrrrrr}
\toprule
Model & Accuracy & Macro-F1 & Trade recall
      & No-trade recall & Trade precision \\
\midrule
Gemini-3-Flash & 68.60 & 65.34 & 76.55 & 65.99 & 42.57 \\
Gemini-3.5-Flash       & 75.06 & 68.94 & 61.89 & 79.40 & 49.74 \\
GPT-4o-mini            & 76.11 & 69.32 & 58.63 & 81.87 & 51.58 \\
GPT-5.6-Luna           & 78.05 & 71.23 & 59.28 & 84.23 & 55.32 \\
GPT-5.6-Sol            & 79.18 & 66.24 & 34.85 & 93.78 & 64.85 \\
\midrule
Always no-trade        & 75.22 & 42.93 & 0.00 & 100.00 & -- \\
Recent-activity persistence
                       & 79.82 & 71.45 & 51.79 & 89.06 & 60.92 \\
\bottomrule
\end{tabular}
\end{table*}

\begin{figure*}[t]
\centering
\includegraphics[width=.98\textwidth]{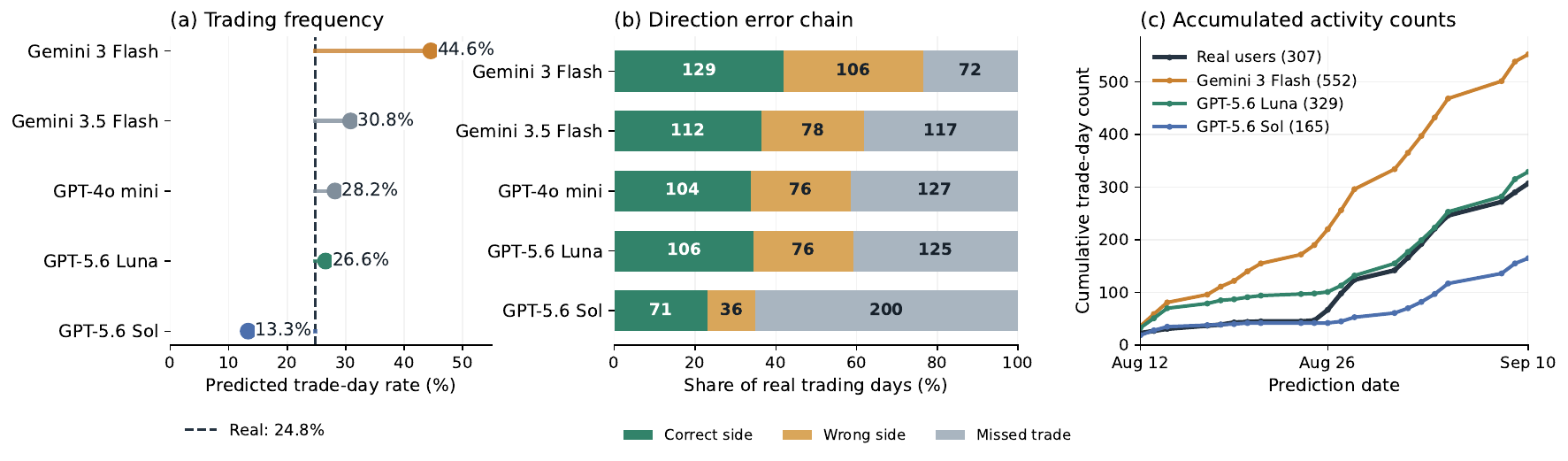}
\caption{Structural mismatch between real and simulated financial decisions.
Simulators exhibit distinct activity biases, and errors propagate from missed
trades to incorrect side composition. Matching cumulative trading frequency
does not imply matching the timing or structure of the underlying decisions.}
\label{fig:structural-bias}
\end{figure*}

\textbf{Matching aggregate trading frequency does not imply temporal fidelity.}
Figure~\ref{fig:structural-bias}(a,c) makes the distinction visible. Some
simulators systematically overpredict user activity, while others underpredict
it. GPT-5.6-Luna is particularly informative: its overall predicted trade rate
is close to the real-user rate, yet it still disagrees with the observed action
on many aligned dates. Consequently, matching total activity can arise even
when individual decisions occur at the wrong times.

The cumulative curves reinforce the same point. Deviations compound in
different directions across simulators, but even a curve that tracks the
observed total does not establish that the same days were predicted correctly.
Aggregate turnover agreement is therefore a weak proxy for temporal behavioral
fidelity.

\textbf{Trade detection does not imply recovery of the action structure.}
Real trading behavior is also more complex than a binary buy-versus-sell
decision. Among the 307 confirmed trading days, 109 are buy-only, 71 are
sell-only, and 127 contain both purchases and sales. Thus, simultaneous buying
and selling accounts for 41.4\% of observed trading days and is a substantial
behavioral mode rather than an edge case. We consequently treat
\texttt{buy\_and\_sell} as a distinct side class.

Recovering this structure remains difficult. The strongest simulator correctly
predicts the three-way side on 129 of the 307 trading days (42.02\%), only
slightly above the 41.37\% majority baseline obtained by always predicting
\texttt{buy\_and\_sell}. Its remaining errors arise through two different
paths: some trades are missed entirely at the activity stage, while others are
detected but assigned the wrong side composition
(Figure~\ref{fig:structural-bias}(b)).

The errors are also asymmetric across action types. Sell-only days are
particularly difficult to recover, with end-to-end recall between 4.23\% and
8.45\% across the five models, whereas recall for
\texttt{buy\_and\_sell} ranges from 41.73\% to 67.72\%. For the joint class,
Gemini-3.5-Flash achieves the strongest F1 (0.647), while GPT-5.6-Sol attains
higher precision but substantially lower recall. This contrast indicates that
models differ not only in overall side accuracy, but also in how readily they
infer multi-action trading days.

\textbf{Recognizing the action structure does not imply recovering its assets.}
We next evaluate asset selection conditionally on the observed side. A trading
day is counted once in side classification, whereas a
\texttt{buy\_and\_sell} day contributes to both the buy-ticker and sell-ticker
retrieval tasks. The resulting evaluation sets contain 236 gold buying days and
198 gold selling days, with the overlap induced by the 127 joint
buy-and-sell days.

Because the simulator contract returns at most five tickers per side, we report
Hit@5 and NDCG@5 in the main text. Table~\ref{tab:ticker_retrieval} reports
candidate-restricted rankings, keeping the scored predictions consistent with
the candidate-generation contract.

\begin{table}[t]
\centering
\footnotesize
\setlength{\tabcolsep}{3pt}
\caption{Conditional ticker retrieval after restricting model outputs to the
candidate pool. Joint buy-and-sell days contribute to both side-specific
evaluation sets.}
\label{tab:ticker_retrieval}
\begin{tabular}{lrrrr}
\toprule
Model &
\shortstack{Buy\\Hit@5} &
\shortstack{Buy\\NDCG@5} &
\shortstack{Sell\\Hit@5} &
\shortstack{Sell\\NDCG@5} \\
\midrule
Gemini-3-Flash
& \textbf{40.7} & 0.259 & \textbf{42.9} & \textbf{0.272} \\
Gemini-3.5-Flash
& 39.0 & \textbf{0.260} & \textbf{42.9} & 0.265 \\
GPT-4o-mini
& 37.7 & 0.247 & 27.8 & 0.164 \\
GPT-5.6-Luna
& 35.2 & 0.223 & \textbf{42.9} & 0.257 \\
GPT-5.6-Sol
& 22.9 & 0.150 & 32.3 & 0.219 \\
\bottomrule
\end{tabular}
\end{table}

Ticker recovery is substantially harder than detecting that a user will trade.
Even the strongest Hit@5 values remain around 40--43\%, meaning that a model
can correctly infer the occurrence or structure of a trade while failing to
recover the assets that constitute the decision. The 127 joint
buy-and-sell days make this separation especially clear: recognizing that a
user will both buy and sell is a different problem from reconstructing the two
corresponding ticker sets.

\textbf{Candidate reachability is a separate bottleneck from asset selection.}
Ticker-level performance also depends on whether the relevant asset is
available to the simulator in the first place. At least one true bought ticker
appears in the candidate pool on only 73.31\% of buying days, compared with
91.92\% of selling days. Moreover, 116 of the 236 buying days contain at least
one gold ticker that is absent from the candidate pool.

Consequently, a ticker miss can arise before the ranking stage: the correct
asset may never have been reachable. At the same time, imperfect candidate
coverage does not explain the entire gap. Candidate-restricted performance
remains far below the corresponding candidate oracle, indicating substantial
residual error even when useful assets are available. The end-to-end ticker
metric therefore reflects at least two distinct failure modes: candidate
generation and downstream asset selection.

This distinction is important for interpreting simulator fidelity. A low ticker
score does not necessarily imply that the model ignored an available asset, but
a high candidate recall would also not guarantee that the simulator could
select the correct asset once exposed to it.

\textbf{Behavioral agreement does not guarantee economic fidelity.}
The previous analyses evaluate discrete behavioral decisions. We finally ask
whether behavioral similarity translates into similar portfolio trajectories.
We select three high-fidelity users with sufficient observation coverage,
trading activity, and end-of-day holdings changes to support a meaningful
comparison.

For each user, the simulated portfolio is initialized as an exact copy of the
real portfolio at the first prediction cutoff. The simulated portfolio is then
updated using only the simulator's predicted buy and sell tickers. Trade sizes
are calibrated from that user's historical behavior, and both real and
simulated portfolios are marked using the same adjusted close prices. We
therefore interpret the resulting curves as \emph{decision-attribution
trajectories}, rather than realized P\&L: they isolate the economic
consequences of different simulated decision paths under a common accounting
rule.

\begin{figure*}[t]
\centering
\includegraphics[width=\textwidth]{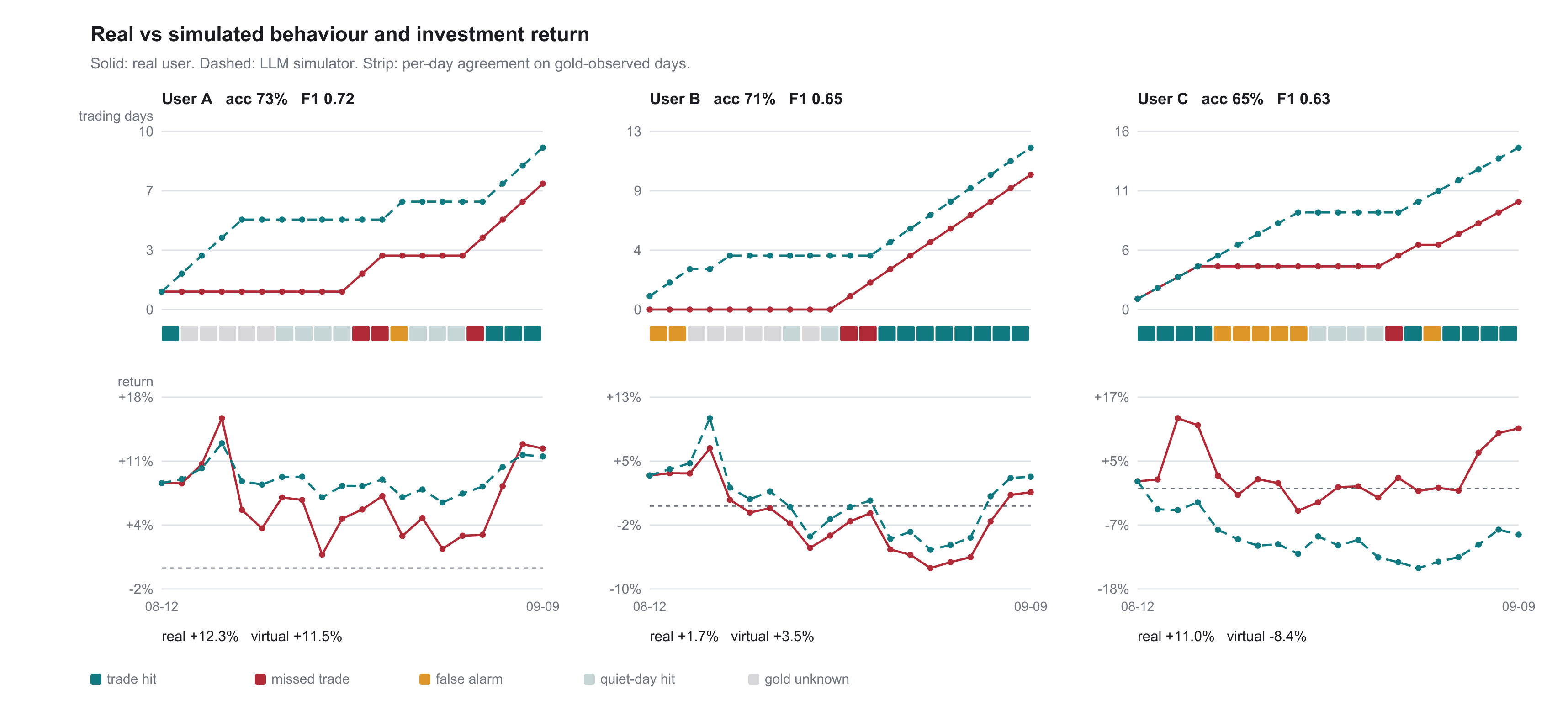}
\caption{Behavioral agreement does not necessarily translate into economic
fidelity. Similar activity predictions can produce similar portfolio paths
(Users A and B), but can also mask substantial economic divergence when
asset-level decisions differ from the real ones (User C).}
\label{fig:rq1-return-curves}
\end{figure*}

Figure~\ref{fig:rq1-return-curves} reveals that activity-level fidelity and
economic fidelity can diverge sharply. For Users A and B, relatively strong
behavioral agreement is accompanied by similar terminal returns: the simulated
portfolios end within 0.82 and 1.86 percentage points of their corresponding
real portfolios. These cases show that the simulator can sometimes recover a
decision path with similar economic consequences.

User C provides the more informative counterexample. Despite an activity
macro-F1 of 0.627, the real portfolio returns +11.0\% while the simulated
portfolio returns $-8.37\%$, producing a 19.37 percentage-point gap. Thus,
moderate agreement on \emph{whether} a user trades can coexist with a
qualitatively different economic trajectory when the simulator does not recover
the correct assets, direction, or position scale.

\textbf{RQ1 takeaway.}
The results expose a hierarchy of financial-simulator fidelity. LLMs can
approximate broad user-level trading propensity, but apparent agreement
progressively weakens as the evaluation moves from \emph{whether} a user trades
to \emph{how} the action is structured, \emph{which assets} are selected, and
\emph{what economic trajectory} those decisions induce.

In particular, strong activity accuracy can largely reflect short-term
behavioral persistence; matching aggregate trading frequency does not imply
correct timing; recognizing a joint buy-and-sell action does not imply
recovering the corresponding assets; and activity-level agreement can still
produce substantially different portfolio outcomes. Financial user simulation
should therefore be evaluated as a hierarchy of behavioral decisions rather
than as a single next-action prediction problem.

\subsection{RQ2: Does the Simulator Recover User Investment Logic?}
\label{sec:rq2}

We investigate whether an LLM-based user simulator recovers the logic underlying
a user's investment decisions. We separate two questions that are easily
conflated. First, which behavioral evidence actually governs the simulator's
next-action prediction? Second, can an observable pre-trade behavior be
identified as a causal trigger of the user's real trading decision? The first
concerns the decision logic learned by the simulator, whereas the second
concerns the behavioral mechanism of the user. An input may causally change the
simulator's prediction without being a cause of the user's behavior.

\textbf{Recent trading history dominates the simulator's activity decision.}
We first intervene directly on the simulator's information environment through
paired prompt ablations with \texttt{gpt-5.6-luna}. For each user--date pair,
we hold the model, target date, prompt template, and decoding configuration
fixed, while replacing one evidence block with an explicit \emph{withheld}
marker. We separately remove recent trade records, holdings and portfolio
changes, seven-day app behavior, market context, and the long-horizon profile.

The resulting dependence is strongly asymmetric. Removing recent trade records
reduces activity accuracy by 23.49 percentage points
(95\% CI: $[-33.00,-13.65]$), while no other evidence block produces a
reliable loss in activity accuracy. This suggests that the simulator's
trade/no-trade decision is anchored primarily in \emph{behavioral persistence}:
recent trading activity provides a strong prior that the user will remain
active. Holdings, app interactions, market conditions, and the long-horizon
profile contribute comparatively little to the binary decision of whether a
trade occurs.

\begin{table*}[t]
\centering
\small
\setlength{\tabcolsep}{4pt}
\caption{Paired evidence ablation for simulator decisions. Removing recent
trade records uniquely produces a reliable loss in activity accuracy, whereas
asset-level predictions remain sensitive to nearly every evidence source.
Negative $\Delta$ accuracy indicates worse activity prediction relative to the
paired full-information control.}
\label{tab:rq2-ablation}
\begin{tabular}{lrrrrrr}
\toprule
Withheld evidence &
\shortstack{$\Delta$ accuracy\\(pp)} &
95\% CI &
\shortstack{Activity\\flip} &
\shortstack{Side\\flip} &
\shortstack{Ticker-list\\flip} &
\shortstack{Confidence\\decrease} \\
\midrule
Recent trade records
& \textbf{-23.49}
& \textbf{[-33.00, -13.65]}
& \textbf{36.5\%}
& \textbf{27.9\%}
& \textbf{76.6\%}
& \textbf{57.0\%} \\
Holdings and portfolio changes
& -1.40
& [-3.98, 1.15]
& 7.0\%
& 16.2\%
& 58.6\%
& 8.8\% \\
7-day app behavior
& -0.23
& [-2.16, 1.41]
& 4.4\%
& 9.9\%
& 56.8\%
& 8.1\% \\
Market context
& -1.16
& [-3.20, 0.74]
& 4.0\%
& 9.9\%
& 57.7\%
& 10.2\% \\
Long-horizon profile
& -0.23
& [-2.88, 2.14]
& 6.7\%
& 12.6\%
& 60.4\%
& 8.4\% \\
\bottomrule
\end{tabular}
\end{table*}

\textbf{The simulator is more stable about when to trade than what to trade.}
A different pattern emerges at the asset level. Withholding any evidence block
changes more than half of the predicted ticker lists on gold-observed trade
days, even when the model's trade/no-trade accuracy remains nearly unchanged.
The simulator can therefore preserve its belief that a user will trade while
substantially revising which asset the user is expected to select.

This separation suggests that activity timing and asset selection rely on
different forms of inferred user logic. The former is dominated by a relatively
stable short-term activity prior, whereas the latter appears to be assembled
from several partially redundant signals. The simulator therefore captures
short-term behavioral momentum more robustly than asset-specific investment
logic. We do not interpret the ablation effects additively, since the evidence
blocks overlap: portfolio changes may partially reveal recent trading, and
holdings and market context contain related portfolio information.

\begin{figure}[t]
\centering
\includegraphics[width=\linewidth]{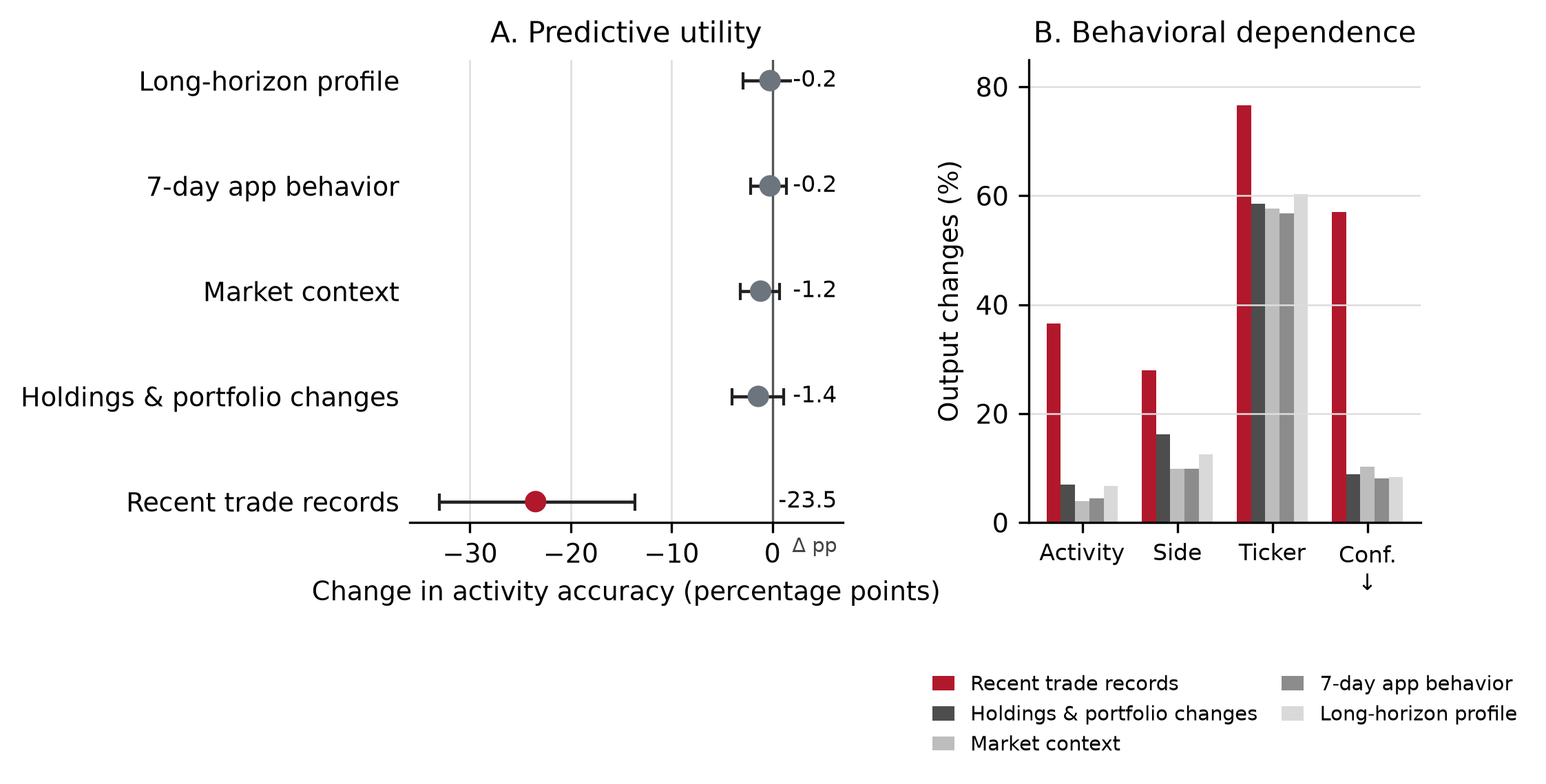}
\caption{Evidence dependence of simulator decisions. Recent trade history is
uniquely important for determining whether a user will trade, whereas predicted
ticker lists remain sensitive to nearly every evidence source. The contrast
suggests that the simulator captures short-term activity momentum more reliably
than asset-specific investment logic.}
\label{fig:rq2-ablation}
\end{figure}

\textbf{Research escalation marks an approaching trade, but is not identified
as its cause.}
We next move from the simulator's internal dependence to the user's observed
behavior. We ask whether a concrete pre-trade action---an escalation in
ticker-specific research---can be interpreted as a trigger of subsequent
trading. We pre-specify an observational target-trial emulation at the
user--ticker--date level. Treatment is defined as a one-day escalation into
research following prior quote attention and two no-research baseline sessions,
and the outcome is an executed same-ticker trade on the next trading day.

The opportunity set is fixed before treatment and contains held, viewed, or
recently traded tickers. We adjust for user, ticker, and date fixed effects
together with pre-treatment holdings, trading history, attention, portfolio
state, and available market features using IPW/MSM and AIPW estimators. The
primary risk set contains 15,125 opportunities, including 395 treated
opportunities from 80 users over 21 dates.

Research escalation is positively associated with imminent trading. The
adjusted probability of a next-day same-ticker trade increases by 3.94
percentage points, with similar positive associations for both buys and sells.
This makes intensified ticker-specific research a useful behavioral marker of
an approaching investment decision.

The identification diagnostics, however, rule out a causal interpretation.
Although common support is largely retained, post-weighting covariate balance
remains inadequate, with a maximum weighted standardized mean difference of
0.324 relative to the pre-specified 0.10 threshold. More importantly, the
future-behavior placebo remains positive (+5.12 pp; 95\% CI:
$[1.58,10.07]$). These failures are consistent with residual time-varying
confounding.

\begin{figure*}[t]
\centering
\includegraphics[width=.96\textwidth]{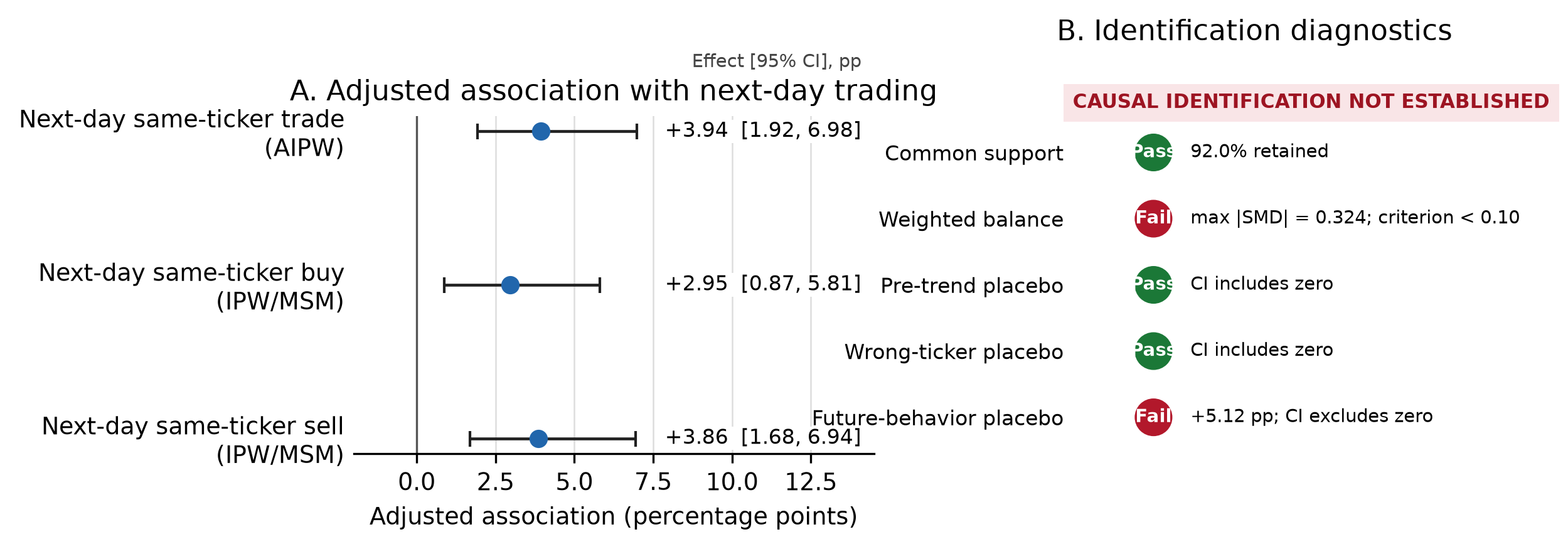}
\caption{Research escalation is positively associated with next-day
same-ticker trading, but the identification diagnostics do not support a causal
interpretation. Residual covariate imbalance and a positive future-behavior
placebo indicate that intensified research is better viewed as a marker of an
emerging trading decision than as an identified trigger of that decision.}
\label{fig:rq2-target-trial}
\end{figure*}

A plausible explanation is that an unobserved increase in ticker-specific
investment intent jointly produces both additional research and the subsequent
trade. Under this mechanism,

\[
\text{latent investment intent}
\longrightarrow
\left\{
\begin{array}{l}
\text{research escalation},\\
\text{subsequent trade}.
\end{array}
\right.
\]

The observed temporal sequence of research followed by trading therefore does
not establish

\[
\text{research escalation}
\longrightarrow
\text{trade}.
\]

Instead, research activity may be an observable manifestation of a decision
process that is already underway. We consequently treat research escalation as
a temporally proximal behavioral marker rather than an identified causal
trigger.

\textbf{The simulator recovers predictive logic, not a complete causal
decision mechanism.}
Taken together, the two experiments expose different levels of investment
logic. The prompt intervention shows that recent trading history causally
governs the simulator's activity prediction, providing evidence that the model
has learned a meaningful short-term behavioral dependency. However, its
ticker-level predictions are substantially more fragile, suggesting that the
model does not recover an equally stable asset-selection mechanism.

The observational analysis reveals a complementary limitation on interpreting
the user's behavior. Ticker-specific research contains information about
imminent trading, but the available data do not establish that research itself
causes the decision. We therefore distinguish three increasingly demanding
notions of behavioral understanding: \emph{predictive dependence}, where an
evidence source changes the simulator's prediction; \emph{behavioral
association}, where an observable behavior precedes and predicts real trading;
and \emph{causal mechanism}, where intervening on that behavior would change
the user's decision. Our results provide evidence for the first two, but not
the third.

Thus, the simulator demonstrates partial understanding of investment logic:
it captures \emph{when} recent behavioral momentum makes trading more likely,
but it is less reliable in recovering \emph{what} the user will trade, and the
observational data are insufficient to establish \emph{why} the user ultimately
acts.

\section{Related Work}
\subsection{LLM as Human Simulator}
Advances in large language models (LLMs) have enabled increasingly realistic simulation of human cognition and interaction across diverse domains, including dialogue~\cite{chan2023chateval}, recommendation~\cite{chen2026realworldhumanbehaviorsimulation,10.1145/3708985,10.1145/3589334.3645537}, and autonomous driving~\cite{jin2024surrealdriverdesigningllmpoweredgenerative}. Seminal works such as Generative Agents~\cite{10.1145/3586183.3606763} and BASES~\cite{ren-etal-2024-bases} further extend this paradigm to long-horizon social behavior and web search. However, most existing user simulators are evaluated in controlled environments, synthetic sandboxes, or relatively stable interaction settings, leaving their effectiveness in financial decision making largely unexplored. In particular, prior work does not examine whether LLM-based simulators can infer an investor's latent decision process and subsequent trading action from heterogeneous behavioral signals under dynamically evolving market conditions. This gap motivates benchmarks grounded in longitudinal platform behavior and point-in-time market information. To this end, we introduce a pilot financial benchmark that jointly captures long-horizon paper-trading behavior and market dynamics, enabling an initial evaluation of LLM-based simulators in modeling investor decision processes and predicting next-step investment actions.

\subsection{Financial Agent Application}
Financial user simulation has evolved from rule-based dialogue models to LLM-based investor agents. Early work introduced user simulators for personalized dialogue management in financial product recommendation~\cite{10.1145/3350546.3352501}. More recent studies, such as StockAgent~\cite{repec:arx:papers:2407.18957} and TwinMarket~\cite{NEURIPS2025_5bf234ec}, employ LLM-based agents to simulate investor trading behavior under dynamic market conditions, with TwinMarket further examining how individual trading decisions give rise to collective market dynamics. Conv-FinRe~\cite{10.1145/3805712.3808658} constructs longitudinal financial recommendation scenarios from real market data and human decision trajectories, highlighting the potential discrepancy between observed investor behavior and underlying utility. More recently, ShiJianBench~\cite{gong2026shijianbenchdialoguedecisionlonghorizon} introduces investor simulators with evolving internal states to study how financial recommendations influence users' subsequent investment decisions over long horizons. However, existing studies have not systematically investigated whether user simulators reproduce individual, heterogeneous behavior across scenarios and over extended time horizons.

\section{Conclusion}

We conducted a longitudinal empirical study of LLMs as financial user simulators, examining whether apparent behavioral similarity reflects faithful reproduction of individual decision-making. Our evaluation separates fidelity across trade occurrence, action structure, asset selection, and downstream portfolio consequences, and further distinguishes predictive performance from the evidence dependence underlying model decisions.

The results reveal a consistent hierarchy of behavioral fidelity. Current LLMs can capture aspects of coarse trading propensity, but a simple recent-activity persistence rule remains a strong baseline for predicting whether an individual trades. Agreement deteriorates further when recovering buy--sell structure and traded assets, and similar activity-level predictions can produce substantially different portfolio trajectories. These findings show that aggregate or coarse action agreement can substantially overstate individual-level simulation fidelity.

Our evidence analysis further suggests that current simulators recover predictive regularities more reliably than stable decision mechanisms. Recent trading history strongly influences activity prediction, whereas asset selection is substantially more sensitive to changes in the available evidence. Although intensified ticker-specific research is associated with imminent trading in the observed participant trajectories, the available data do not support interpreting this relationship causally.

Taken together, these results suggest that evaluating LLM user simulators requires more than measuring whether their outputs resemble observed actions. Behavioral fidelity should be assessed across temporal alignment, decision structure, object-level choices, and downstream consequences, while predictive dependence should be distinguished from evidence about the causal mechanisms underlying human behavior.


\bibliography{custom}

\appendix

\section{Experimental Setting}
\label{sec:experiment_setting}

\subsection{Baselines} 
We evaluate a diverse set of representative LLMs, including both proprietary and open-weight models. The proprietary models the GPT series (GPT-5.6 Sol\cite{openai2026gpt56}, GPT-5.6 luna \cite{openai2026gpt56} and GPT-5.4o-mini\cite{openai2024gpt4omini}), and the Geimini series (Geimini-3-Flash\cite{gemini3flash2025google} and Geimini-3.5-Flash\cite{gemini35flash2026})

\begin{comment}
The open-weight models include DeepSeek-V4~\cite{deepseekai2026deepseekv4highlyefficientmilliontoken} and GLM5.3-Flash~\cite{glm5team2026glm5vibecodingagentic}.
\end{comment}

\subsection{DataSet}

\textbf{Data Source}
Using real brokerage records for research can raise substantial privacy, regulatory, and data-governance concerns. We therefore constructed a controlled paper-trading environment and did not access participants' real brokerage accounts, money holdings, credentials, or transaction records.

We recruited 80 volunteer participants. Participants could buy, sell, rebalance, or remain inactive at their discretion under real-time market conditions. Orders were simulated exclusively with non-redeemable virtual
funds and had no effect on participants' real assets, brokerage accounts, or financial positions. We recorded the resulting paper-trading trajectories over a three-month observation period.

\textbf{Privacy Protection}
Although the platform did not collect real brokerage data, the resulting interaction logs constitute participant-generated behavioral data. We therefore de-identify all data used for analysis and release. A locally deployed
DeepSeek-V4-Flash model~\cite{deepseekai2026deepseekv4highlyefficientmilliontoken} assists in detecting personally identifiable information in any free-text or account-associated fields. Detected identifiers are replaced with
standardized placeholders (e.g., \texttt{[NAME]} and \texttt{[AGE]}). We then conduct manual quality assurance before releasing the processed dataset. No raw identifiers, real account credentials, money positions, and brokerage transactions are included in the released data.

\textbf{Construction Pipeline}
We collected data from 80 volunteers using the \textit{Paper Trading} platform, comprising 230K+ interactions. The data span multiple financial scenarios, including market-information and news browsing, stock/ETF viewing, simulated trading activity, virtual portfolio states, and interactions with the platform chatbot.

\textbf{Data Processing}
We clean duplicated, incomplete, and noisy records and organize all interactions into unified chronological behavior sequences. For each user, earlier historical behaviors are further used to construct a financial profile capturing relatively stable characteristics such as investment preferences, trading patterns, and portfolio characteristics. For viewed financial news, we clean the original text and use \textbf{DeepSeek-V4}\cite{deepseekai2026deepseekv4highlyefficientmilliontoken} to generate concise summaries, which are inserted into the corresponding positions of the behavior sequence.

\textbf{User Diversity.}
To preserve behavioral heterogeneity, our dataset includes both active and inactive participants with diverse paper-trading styles, including high-risk trading and long-term investing. Participants also vary in financial experience, ranging from experienced market participants to non-professional retail investors.

\section{User Profile Construction}
\label{sec:user_profile}

\begin{figure}[t]
\centering
\includegraphics[width=\linewidth]{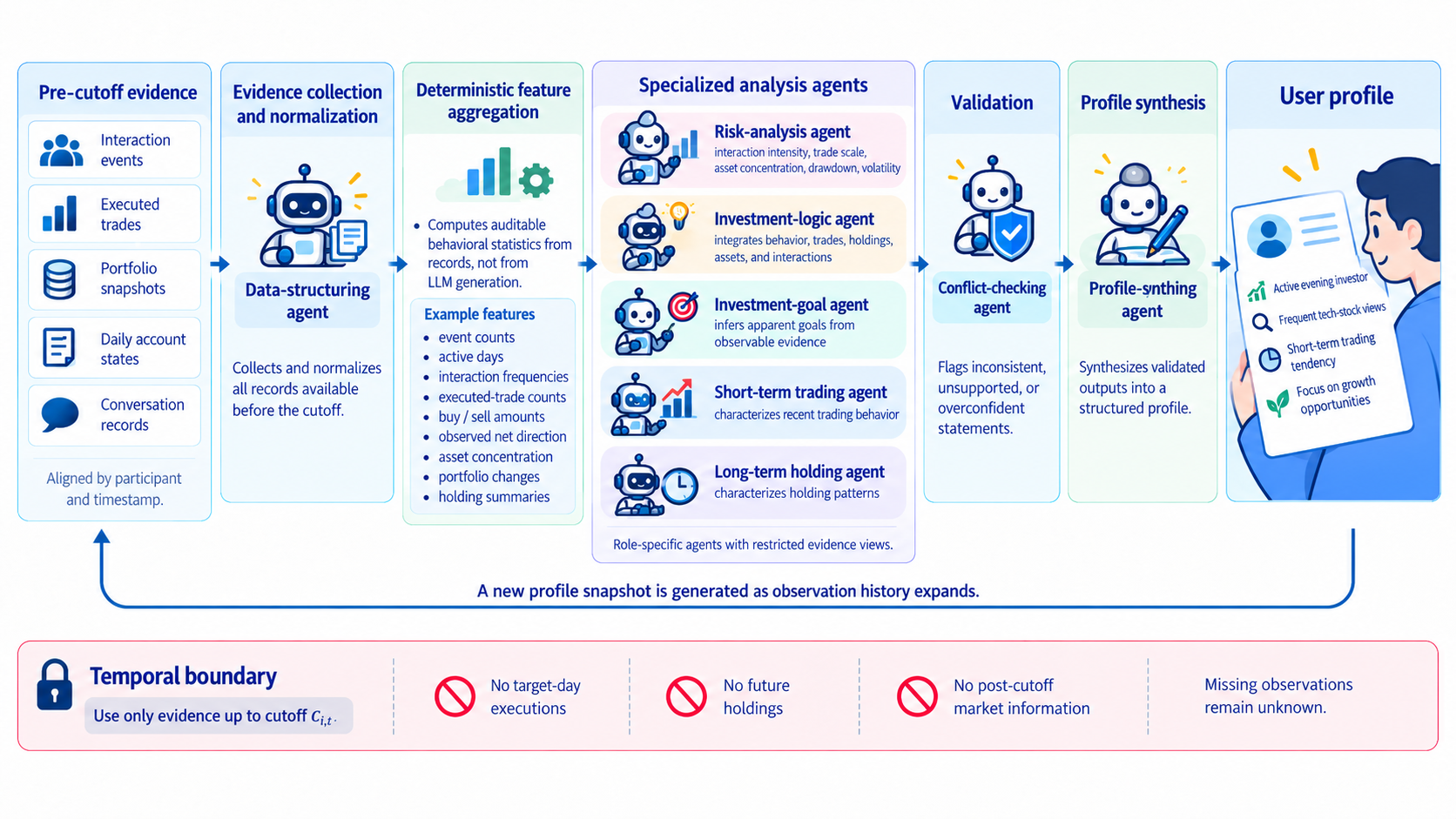}
\caption{Cutoff-safe multi-agent user-profile construction. Raw
pre-cutoff records are normalized and converted into deterministic
features, analyzed by role-specific agents, checked for conflicts, and
synthesized into a versioned base profile.}
\label{fig:profile-pipeline}
\end{figure}

We construct a cutoff-safe user profile with a staged multi-agent pipeline\cite{wang2026llmbaseduserpersonasrecommendations,10.1007/978-3-031-05409-9_40}. The profile is intended to represent a participant's \emph{observable behavioral history}, rather than unobserved psychological traits or ground-truth investment preferences. Figure~\ref{fig:profile-pipeline} summarizes the construction process.

\paragraph{Evidence collection and normalization.}
For each participant and prediction cutoff $c_{i,t}$, a data-structuring agent collects and normalizes all records available before the cutoff. The evidence streams include interaction events (e.g., viewing, checking, reading, searching, and switching views), executed trades, portfolio snapshots, daily account states, and conversation records. Records are aligned by participant and timestamp, and target-day or post-cutoff information is excluded before any downstream agent is called.

\paragraph{Deterministic feature computation.}
A feature-aggregation module converts the normalized records into
auditable behavioral statistics. These include the number of events,
active days, interaction frequencies, executed-trade counts, buy and
sell amounts, observed net direction, asset concentration, portfolio
changes, and holding-related summaries. These quantities are computed directly from the underlying records and are not generated by an LLM.

\paragraph{Specialized analysis agents.}
The structured evidence is then routed to role-specific agents with
restricted evidence views. A risk-analysis agent examines observable
risk-related behavior, including transaction intensity, trade scale,
asset concentration, drawdown, and portfolio volatility. An investment-logic agent integrates cross-source signals from behavior, trades, holdings, assets, and interactions to characterize the participant's apparent investment logic. Additional agents analyze investment goals, short-term trading behavior, and long-term holding patterns. Each agent produces evidence-linked conclusions together with uncertainty or missing-evidence indicators.

\paragraph{Validation and profile synthesis.}
A conflict-checking agent compares the outputs of the specialized
agents and flags inconsistent, unsupported, or overconfident
statements. A profile-writing agent then synthesizes the validated
outputs into a structured profile containing (i) a concise overview,
(ii) dimension-level summaries, (iii) evidence-grounded highlights,
and (iv) explicit caveats. Missing observations remain unknown and are not converted into negative behavioral claims.

\paragraph{Temporal versioning and model-facing representation.}
The profile is versioned by participant and cutoff. As the observation history expands, a new profile snapshot is generated using only evidence available at the new cutoff. The simulator receives the versioned profile together with separate short-horizon behavior, portfolio, and market-context blocks. Thus, the long-horizon profile captures persistent observable patterns, whereas the rolling short-horizon blocks capture recent conditions relevant to the next decision.

\end{document}